\documentclass{article}
\usepackage{ijcai21}

\pdfoutput=1

\usepackage{times}
\usepackage{soul}
\usepackage{url}
\usepackage[hidelinks]{hyperref}
\usepackage[utf8]{inputenc}
\usepackage[small]{caption}
\usepackage{amsmath}
\usepackage{amsthm}
\usepackage{booktabs}
\usepackage{graphicx}
\graphicspath{ {./images/} }

\usepackage{MnSymbol}
\usepackage{xcolor}

\usepackage[inline, shortlabels]{enumitem}
\setlist{nolistsep, leftmargin=*}

\newcommand{\AR}[1]{\textcolor{black}{#1}}
\newcommand{\todo}[1]{}
\newcommand{\del}[1]{}

\newcommand{\ARF}[1]{\textcolor{black}{#1}}
\newcommand{\ARD}[1]{}

\def\wrt{wrt}
\def\resp{resp.}

\def\argexpl{AF-based explanation}
\def\argexpls{AF-based explanations}

\def\Argexpls{AF-based Explanations}
\def\AACBR{AA-CBR}
\def\AALD{AA-LD}

\def\AAALP{AA-ALP}
\def\AAAS{AA-AS}
\def\dialogue{Dialogue}

\newcommand{\tuple}[1]{\langle{#1}\rangle}
\def\Args{\ensuremath{\mathit{Args}}}
\def\Atts{\ensuremath{\mathcal{R}^-}}
\def\Supps{\ensuremath{\mathcal{R}^+}}

\def\arga{\ensuremath{\ensuremath{\alpha}}}
\def\argb{\ensuremath{\ensuremath{\beta}}}

\def\graph{\ensuremath{\mathcal{G}}} 
\def\prop{\ensuremath{\mathsf{P}}} 

\usepackage{hyperref}
\hypersetup{
	pdftitle={Argumentative XAI: A Survey},
	pdfauthor={Kristijonas \v{C}yras, Antonio Rago, Emanuele Albini, Pietro Baroni, Francesca Toni},
	pdfsubject={Argumentation, Explanations, XAI}
}

\title{Argumentative XAI: A Survey\thanks{To appear in IJCAI 2021 Survey Track. Do not cite this preprint.}}

\author{
Kristijonas \v{C}yras$^1$\and
Antonio Rago$^{2}$\and
Emanuele Albini$^2$\and
Pietro Baroni$^3$\And
Francesca Toni$^{2}$\\
\affiliations
$^1$Ericsson Research, Sweden\\
$^2$Department of Computing, Imperial College London, UK\\
$^3$Dipartimento di Ingegneria dell'Informazione, Universit{\`{a}} degli Studi di Brescia, Italy\\
\emails
kristijonas.cyras@ericsson.com,
\{a.rago, emanuele, ft\}@imperial.ac.uk,
pietro.baroni@unibs.it
}

\begin{document}

\maketitle

\begin{abstract}
Explainable AI (XAI) has been investigated for decades and, together with AI itself, has witnessed unprecedented growth in recent years. 
Among various approaches to XAI, argumentative models have been advocated in both the AI and social science literature, as their dialectical nature appears to match some basic desirable features of the explanation activity.
In this survey we overview XAI approaches built using methods from the field of \emph{computational argumentation}, leveraging its wide array of reasoning abstractions and explanation delivery methods. 
We overview the literature focusing on different types of explanation (intrinsic and post-hoc), different models with which argumentation-based explanations are deployed, different forms of delivery, and different argumentation frameworks they use. 
We also lay out a roadmap for future work. 
\end{abstract}


\section{Introduction}\label{sec:Intro}

Explainable AI (XAI) has attracted a great amount of attention in recent years, due mostly to its role in bridging applications of AI and humans who develop or use them. 
Several approaches to support XAI have been proposed (see e.g.\ some recent overviews~\cite{Adadi:Berrada:2018,Guidotti}) and the crucial role of XAI in human-machine settings has been emphasised~\cite{Rosenfeld:Richardson:2019}. 
Whereas several recent efforts in XAI are focused on explaining machine learning models~\cite{Adadi:Berrada:2018}, 
XAI has also been a recurrent concern in other AI settings, e.g.\ expert systems~\cite{Swartout:Paris:Moore:1991}, 
answer set programming~\cite{Fandinno:Schulz:2019} 
and planning~\cite{Chakraborti.et.al:2020}.

We provide a comprehensive survey of literature in XAI viewing explanations as \emph{argumentative} (independently of the underlying methods to be explained). 
Argumentative explanations are advocated in the social sciences (e.g.~\cite{argExpl}), focusing on the human perspective, and
argumentation's potential advantages for XAI have been pointed out by several, e.g.~\cite{Moulin.et.al:2002,Bex_16,Sklar_18}. 

Many methods for generating explanations in XAI can be seen as argumentative. 
Indeed, attribution methods, 
including model-agnostic~\cite{Lundberg_17} and model-specific~\cite{Shih_18} approaches, link inputs to outputs via (weighted) positive and negative relations, and 
contrastive explanations identify reasons pro and con 
outputs~\cite{Chakraborti.et.al:2020,Miller_19}.
In this survey we focus instead on overtly argumentative approaches, with an emphasis on the several existing XAI solutions using forms of \emph{computational argumentation}
(see \cite{Baroni_18} for a recent overview of this field of symbolic AI).

The application of computational argumentation
to XAI is supported by its strong theoretical and algorithmic 
foundations and the flexibility it affords, particularly in the wide variety of 
\emph{argumentation frameworks} (AFs) on offer in the literature. 
These AFs include ways to specify \emph{arguments} and \emph{dialectical relations} between them, as well as \emph{semantics} to evaluate the dialectical  
\emph{acceptability} or \emph{strength} of arguments, 
while differing (sometimes substantially) in how they define these components. 
When AFs are used to obtain explanations,
(weighted) arguments and dialectical relations
may suitably represent anything from 
input data, e.g.\ categorical data 
or pixels in an image, 
to knowledge, e.g.\ rules
, to internal components of the method being explained, e.g.\ filters in convolutional neural networks,
to 
problem formalisations, e.g.\ planning, scheduling or decision making models, 
to outputs, e.g.\ classifications,  recommendations, or logical inference. 
This flexibility and wide-ranging applicability has led to a 
multitude 
of methods for 
\emph{\argexpls{}}, providing the motivation  
and need for this survey. 


After giving 
some background (§\ref{sec:AFs}), our contributions are
:
\begin{itemize}
    \item 
    we overview  the 
    literature on \argexpls{}, cataloguing 
    representative approaches according to what they explain and by means of which AF (§\ref{sec:applications});
    \item we overview the 
    prevalent forms which \argexpls{} take \ARD{when }\ARF{after being }drawn from AFs 
    (§\ref{sec:expl});
    \item 
    we lay out a roadmap for future 
    work, covering: 
    the need to focus on properties of \argexpls{},  computational aspects, and further applications and other potential 
    developments of \argexpls{} 
    (§\ref{sec:roadmap}).
\end{itemize}

We ignore 
argumentative explanations based on informal notions or models
lacking aspects of AFs (notably, semantics) and
application domains of argumentative XAI, covered in a recent, orthogonal survey \cite{Vassiliades:Bassiliades:Patkos:2021}.

\section{Argumentation Frameworks}
\label{sec:AFs}

In this section we give a brief, high-level overview of the argumentation frameworks (AFs)  \emph{used so far to support XAI} in the literature, focusing on how these various AFs understand \emph{arguments}, \emph{dialectical relations} and \emph{semantics}.

\begin{figure}[t!]
\centering
\includegraphics[width=1\linewidth]{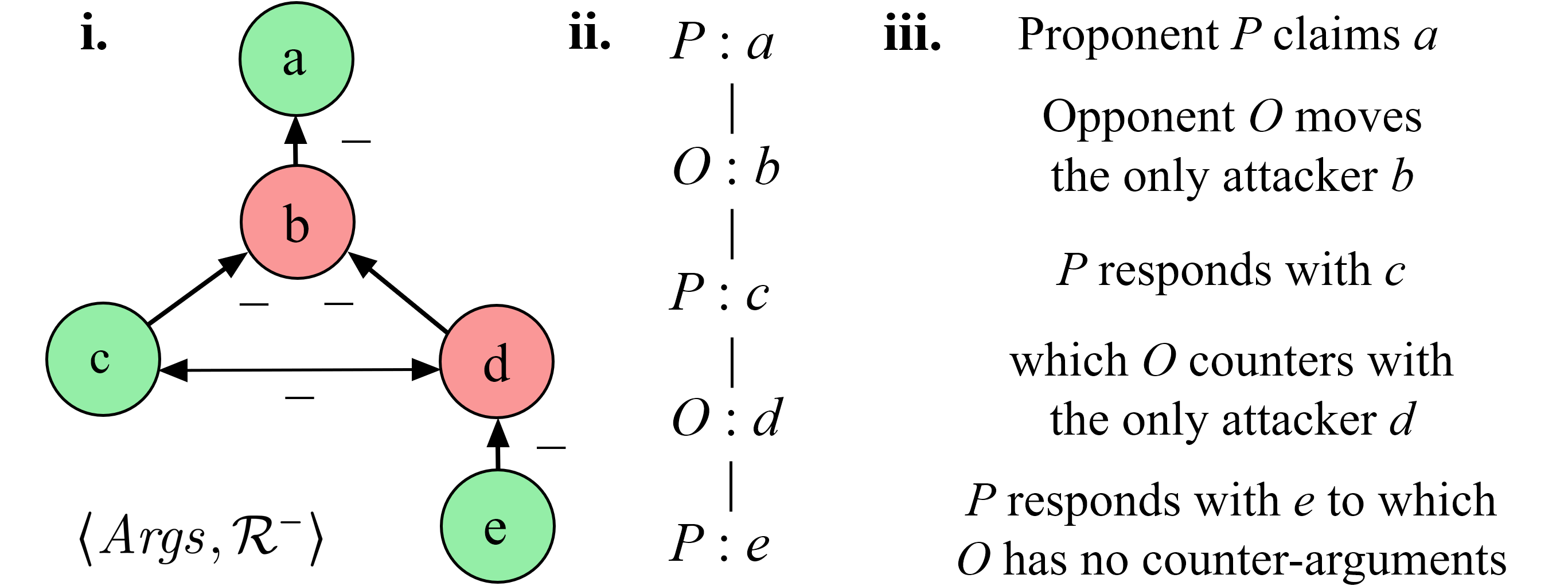}
\caption{(i) An AA framework {$\langle \Args, \Atts \rangle$} visualised as a graph, with nodes as arguments (labelled $a, b, c, d, e$) and directed edges as attacks (labelled $-$). 
Here, the extension $\{a, b\}$ is not conflict-free, but $\{a, c \}$ is.
Arguments (not) in the grounded extension~\protect\cite{Dung_95} are coloured green (resp.\ red). 
(ii) A \emph{grounded dispute tree} (DT) \protect\cite{Dung:Mancarella:Toni:2007} between the proponent $P$ and opponent $O$, for the \emph{topic} argument $a$. 
(Intuitively: $a$ is defended from $b$ by $c$; $e$ defends $c$ from $d$.) 
(iii) A simple \emph{dialogical} explanation where the proponent $P$ wins (drawn from the DT in (ii)).
}
\label{fig:arg}
\end{figure}

A first category of approaches sees 
arguments as \emph{abstract} entities (in some given set $\Args$), starting from 
\textbf{Abstract Argumentation (AA)} \cite{Dung_95}. 
Here, a single dialectical relation of \emph{attack} ($\Atts \subseteq \Args \times \Args $) is considered and semantics \ARD{is }\ARF{are }given in terms of mappings from AA frameworks $\langle \Args, \Atts \rangle$ to sets of so-called \emph{extensions}, each a set of arguments $\mathcal{E} \subseteq \Args$ fulfilling some specified, dialectically meaningful constraints
(e.g.\ \emph{conflict-freeness}, amounting to absence of internal conflicts: $\{ \arga, \argb \in \mathcal{E} | (\arga, \argb) \in \Atts \}=\emptyset$). 
Figure~\ref{fig:arg}i illustrates AA frameworks, conflict-freeness and the widely used \emph{grounded extension semantics}. 
Some of the extension-based semantics can be equivalently understood in terms of \emph{dispute trees} \textbf{(DTs)} \cite{Dung:Mancarella:Toni:2007} (as illustrated in Figure~\ref{fig:arg}ii), which form the basis of several approaches to argumentation-based XAI 
(e.g.\ by providing content for the simple dialogical explanations illustrated in Figure~\ref{fig:arg}iii).

The category of AFs with arguments as abstract entities counts several other approaches.
In \textbf{Bipolar Argumentation (BA)}~\cite{Cayrol:05} an additional dialectical relation of \emph{support} ($\Supps \subseteq \Args \times \Args$) is considered, and a variety of semantics in terms of extensions are given 
(reflecting, in particular, different interpretations of support~\cite{support}).
BA frameworks where the attack relation is empty are also called \textbf{Support Argumentation (SA)} frameworks.  
BA (and thus AA) frameworks can be also equipped with \emph{gradual semantics}, in terms of mappings from frameworks  to value assignments for arguments, representing their \emph{dialectical strength}. 
These values may belong to any given (ordered) set (e.g.\ $[0,1]$) and may be influenced by other factors, e.g. in \textbf{Quantitative Bipolar Argumentation (QBA)}~\cite{Baroni_19}, by values ascribing to arguments their initial, \emph{intrinsic strength}.  
Some AFs consider further dialectical relations, e.g.\ \textbf{Tripolar Argumentation (TA)}~\cite{rec} uses a \emph{neutralising} relation (intuitively pushing arguments' strengths towards the neutral value) and \textbf{Generalised Argumentation (GA)}~\cite{Gabbay:16} can in principle use any number of dialectical relations. 
Further, 
among the
approaches 
with abstract arguments, \textbf{Abstract Dialectical Frameworks (ADFs)}~\cite{ADFs} allow for 
generalised notions of dialectical relations and semantics, specified in terms of user-given \emph{acceptance conditions}. 

A second category of AFs focuses on \emph{structured} arguments instead. 
Among 
those
used for explanation: 
\textbf{ABA}~\cite{ABA97} sees arguments as deductions from assumptions using rules, whereas \textbf{ASPIC+}~\cite{ASPIC+tut} and \textbf{DeLP}~\cite{Garcia_13} draw arguments from 
strict and defeasible rules. 
These approaches, like AA, consider only attacks between arguments.
Further, ABA and ASPIC+, like AA, 
use extension-based semantics, whereas DeLP uses \emph{dialectical trees}, similar in spirit to DTs.

Finally, several approaches relevant to this paper use instances of AA for specific choices of (structured) arguments. 
These may be:
(1)
case-based reasoning (CBR)-inspired, with arguments as past or new cases (where features support a given or open label), as in \cite{Cyras_19_ESA,Cocarascu_20};
%
(2) deductions in abductive logic programming (ALP), as in \cite{Wakaki:Nitta:Sawamura:2009};
(3) logical deductions (LD), possibly built from rules, as in \cite{Arioua:Tamani:Croitoru:2015,Brarda_19,Sendi.et.al:2019}; 
or built from argument schemes (AS), as in \cite{Sassoon_19}.
We refer to AA instances using arguments of these forms, \resp, as \textbf{AA-CBR},
\textbf{AA-ALP}, 
\textbf{AA-LD}, and \textbf{AA-AS} frameworks.


\section{Types of Argumentative Explanations}
\label{sec:applications}

Here, we review the literature for models 
explained using argumentative explanations \emph{built from AFs}
, referred to as \emph{\argexpl s}. 
We divide 
them 
into those which are: 
\begin{itemize}
    \item \emph{intrinsic}, i.e.\ defined for models that are natively using argumentative techniques, 
    covered in §\ref{sec:sub:ex-ante};
    \item \emph{post-hoc}, i.e.\
    obtained from non-argumentative models; 
    we further divide the post-hoc 
    \argexpl s depending on whether they
        provide a \emph{complete} (§\ref{sec:sub:post-hoc-match}) or \emph{approximate} (§\ref{sec:sub:post-hoc-approximate}) representation of the explained model.
\end{itemize}

Note that this three-fold 
distinction among types of \argexpl\ is not crisp, and, as we will see, some of the  approaches we survey may be deemed to 
be hybrids.

We use the term `model' in a very general sense, in the spirit of \cite{Geffner:2018}, to stand for 
a variety of systems, amounting, in our review, to the following categories: recommender systems, 
classifiers and probabilistic methods, decision-making and knowledge-based systems, planners and schedulers, as well as tools for logic programming.

Note also that our focus in this section 
is on explaining models \emph{other than} the argumentation process itself,  while  we will overview later, in 
§\ref{sec:expl}, the forms of (intrinsic) \argexpl s adopted by some approaches to explain argumentation/argumentative agents.
Table \ref{table:applications} provides an overview of the surveyed approaches, 
distinguished according to the category of model for which they provide
intrinsic, complete or approximate types of \argexpl s, and indicating the form of AF from which the explanations are obtained.
We exemplify each of our chosen categories of models with selected works.

\begin{table}[t]
\begin{tabular}{lccc}
& 
\!\!\!\!\textbf{Intrinsic
}\!\!\!\! & 
\!\!\!\!\!\!\textbf{Compl
.}\!\!\!\!\!\! & 
\!\!\textbf{Approx.}\!\! \\ 
\hline
\textbf{Recommender Systems}
 & 
 & 
 & 
 \\
 \cite{Briguez_14}
 & 
\!\!DeLP\!\! & 
 & 
 \\
\cite{Rodriguez_17}
& 
\!\!DeLP\!\! &
 & 
 \\
\cite{rec}
&    
 & 
 & 
\!\!TA\!\! \\
\cite{RT}
& 
\!\!\!\!QBA\!\!\!\! & 
 & 
 \\
\cite{Rago_20}
&
 & 
 & 
 \!\!BA\!\! \\ 
\hline
\textbf{Classification} 
& 
&  
&                       
\\
\cite{Cyras_19_ESA}
&
\!\!\!\!\AACBR\!\!\!\! &
&                     
\\
\cite{Sendi.et.al:2019}
 & 
 &
 &                     
 \!\!\AALD\!\! \\ 
\cite{Cocarascu_20} &
\!\!\AACBR\!\!
 &
 & \\
 \cite{argflow} & & & GA\\
\hline
\textbf{Probabilistic Methods}
&
&
&                       
\\
\cite{Timmer_17}
&
&
&
\!\!SA
\!\! \\ 
\cite{Albini_20}
&
&
\!\!\!\!QBA\!\!\!\! & 
\\
\hline
\textbf{Decision Making} &
&
&
\\
\cite{Amgoud_09}
\!\!\!\! &
\!\!AA\!\! &
 &
 \\ 
\cite{Zeng_18}
&
 &
\!\!ABA\!\! &
\\
\cite{Brarda_19}
& 
\!\!\AALD\!\! &
 &
\\
\cite{Zhong_19}
&
 &
\!\!ABA\!\! &
\\
\hline
\textbf{Knowledge-Based Systems}\!\!
&
&
&
\\
\cite{Arioua:Tamani:Croitoru:2015} &
 &
\!\!\AALD\!\!
&
\\
\cite{Kokciyan_20}
&
\!\!\!\!\!\!AA, ASPIC+\!\!\!\!\!\! &
 &
 \\
\hline
\textbf{Planning \& Scheduling}
&
&
&
\\
\cite{Fan_18} &
 &
\!\!ABA\!\! &
 \\
\cite{Cyras_19}
&
&
\!\!AA\!\!
&
 \\
\cite{Collins:Magazzeni:Parsons:2019}
&
&
&
\!\!ASPIC+\!\! \\
\cite{Oren_20}
&
\!\!ASPIC+\!\! &
 &
 \\
\hline
\textbf{Logic Programming} &
&
&
\\
\cite{Wakaki:Nitta:Sawamura:2009} &
& \!\!\!\!AA-ALP\!\!\!\!
& \\
\cite{Schulz:Toni:2016}
&
&
\!\!ABA\!\! &
\\
\cite{Rolf_19} &
& 
\!\!ADF\!\! &
\\
\hline
\end{tabular}
\caption{Overview of 
\argexpl\ approaches (divided 
\wrt\ class of model, 
category of explanation, and 
form of AF used). 
}
\label{table:applications}
\end{table}



\subsection{
Intrinsic Approaches
}
\label{sec:sub:ex-ante}

\begin{figure}[t!]
\centering
\includegraphics[width=1\linewidth]{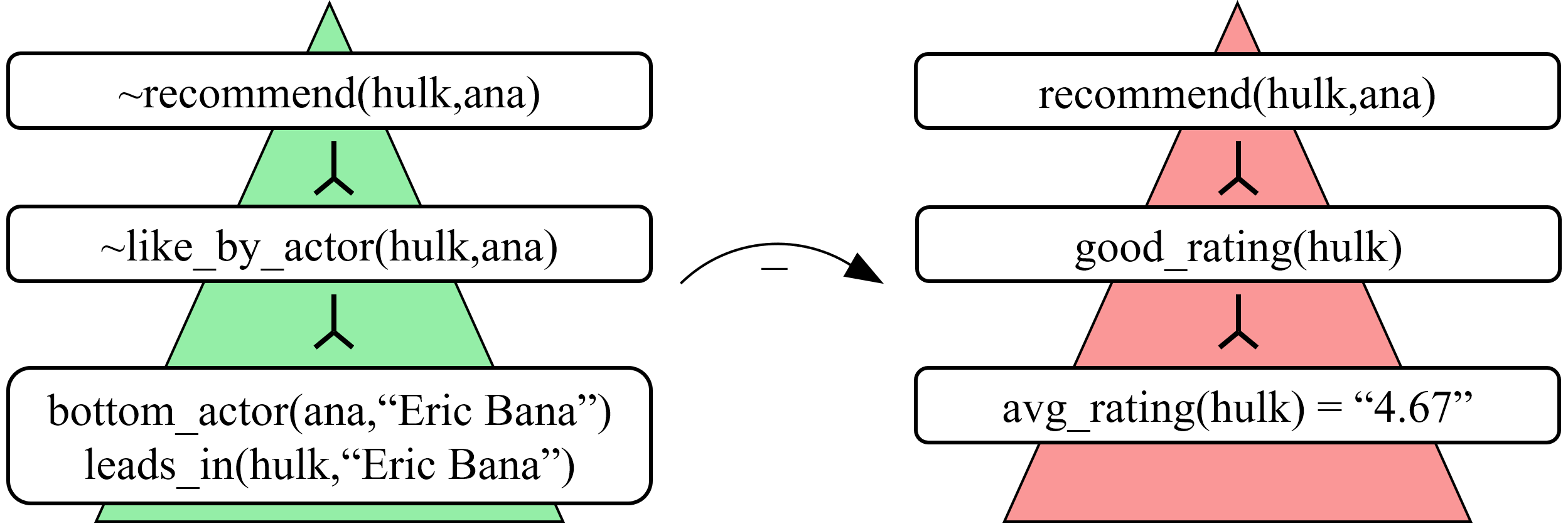}
\caption{Intrinsic \argexpl\ for 
the RS of
\protect\cite{Briguez_14} 
explaining
why the movie \emph{Hulk} was not recommended to the user \emph{Ana}. The undefeated argument against this recommendation (left) attacks (labelled $-$) the argument for the recommendation (right), which is thus defeated. 
DeLP structured arguments are constructed of statements linked by defeasible rules (indicated by $\upY$).}
\label{fig:DeLP}
\end{figure}

One 
application   where intrinsic \argexpl s 
are popular is \textbf{recommender systems} (RSs).
Various RSs have been built with 
DeLP as the main recommendation and explanation engine. 
One is that of \cite{Briguez_14} for the movie domain (see Figure \ref{fig:DeLP})
, handling incomplete and contradictory information and using a comparison criterion to solve conflicting situations
. 
Another is introduced by \cite{Rodriguez_17}, deploying DeLP to provide a hybrid 
RS in an educational setting
, using argumentation to differentiate between different techniques for generating recommendations. 
Other approaches 
deploy 
other forms of AFs. For example, 
\cite{RT} 
provide
argumentation-based review aggregations for movie recommendations and 
conversational explanations extracted from QBA frameworks, in turn extracted from reviews by natural language processing
. 

\cite{Cyras_19_ESA} propose a case-based reasoning (CBR)-inspired but AA-driven \textbf{classification} method for classifying 
and explaining legislation outcomes. 
\cite{Cocarascu_20} \ARD{uses }\ARF{use }a similarly-inspired method for classifying and explaining with a variety of (binary) classification tasks.
In all these settings, classification results from the semantics of AA frameworks constructed from cases and explanations are based on DTs. Whereas in \cite{Cyras_19_ESA} cases amount to categorical data, in \cite{Cocarascu_20} they may be unstructured (e.g. textual).

\cite{Amgoud_09} define a method for argumentative \textbf{decision making}, where AA is 
used to evaluate the acceptance of arguments for and against each potential decision, and decisions are 
selected based on their arguments' acceptance
, with the AA frameworks constituting the explanations.
The AA-based, multi-criteria decision support system of \cite{Brarda_19} 
operates 
similarly but 
 using an instance of AA, whereby arguments are obtained from 
 conditional preference rules
 . 
%
\cite{Kokciyan_20} use AA and argument schemes mappable to ASPIC+ 
to produce 
domain-specific
explanations for \textbf{knowledge-based systems}. 
%
To conclude this section, \cite{Oren_20} devise an explainable system for argument-based \textbf{planning}
, using a variant of ASPIC+ to translate domain rules from a knowledge base to arguments which 
feed into explanations in dialogical format.





\subsection{Complete Post-Hoc Approaches
}
\label{sec:sub:post-hoc-match}

\begin{figure}[t!]
\centering
\includegraphics[width=0.93\linewidth]{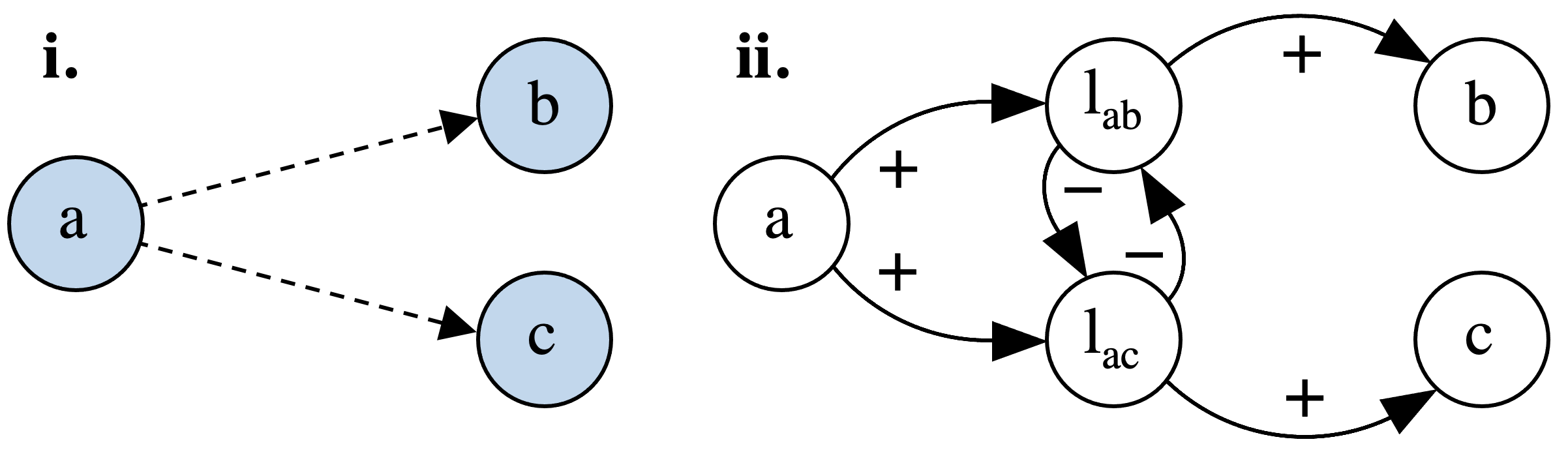}
\caption{Complete post-hoc explanations for PageRank \protect\cite{Albini_20}. (i) Graphical representation of three webpages as nodes, with edges 
showing (hyper)links between them. (ii) Extracted QBA 
framework
, where arguments amount to webpages ($a$, $b$, $c$)  and links $l_{xy}$ (from 
$x$ to 
$y$)
, the strength of an argument (omitted) 
is computed by the PageRank algorithm, and attacks (labelled~$-$) represent negative effects while supports (labelled~$+$) represent positive effects on this strength. It can be seen that $a$'s outgoing links `compete', thus explaining how its PageRank score 
gets distributed to $b$ and $c$.
}
\label{fig:PageRank}
\end{figure}

Complete post-hoc \argexpl s are also deployed in a variety of settings, for several AFs. 
%
With regards to \textbf{probabilistic methods}, \cite{Albini_20} represent web graphs, with webpages as nodes and (hyper)links as edges, with equivalent QBA frameworks in order to explain  
PageRank
scores 
\cite{PageRank} argumentatively (see Figure \ref{fig:PageRank}).

 A popular area for post-hoc \argexpl s is \textbf{decision making}, 
 where ABA frameworks have been shown to 
 represent and explain 
 decision models.
 \cite{Zeng_18} model decision problems as 
 decision graphs with context
 before mapping them into ABA
 , where 
 `best' decisions  correspond to a specific type of (
 admissible) extensions.
\cite{Zhong_19} show that decision problems can be mapped to equivalent ABA frameworks, such that minimally redundant decisions 
correspond to (admissible) 
extensions.
 
 In  \textbf{knowledge-based systems}, 
 \cite{Arioua:Tamani:Croitoru:2015} provide explanations 
 for the purpose of query answering in Datalog under inconsistency-tolerant semantics. 
 Here, entailment of a query from a knowledge base is mapped in a sound and complete fashion onto argument acceptance in AA frameworks constructed from the rules of the knowledge base. 
 Intuitively, argumentative explanations for query entailment/failure 
 result from \emph{defence trees} (similar to DTs).
 
\begin{figure}[t!]
\centering
\includegraphics[width=1\linewidth]{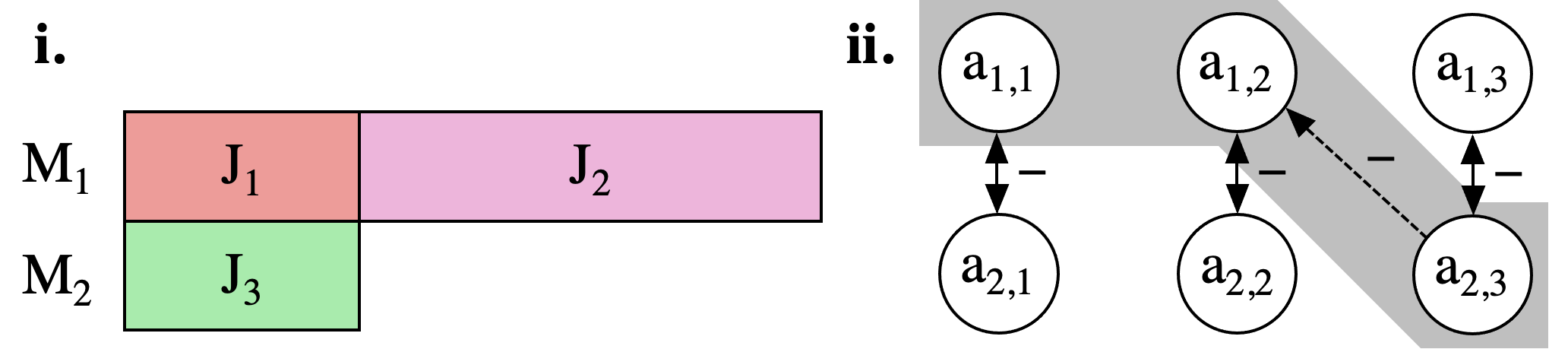}
\caption{Complete post-hoc \argexpl s for scheduling~\protect\cite{Cyras_19}.
(i) An \underline{in}efficient schedule 
with jobs $J_1$, $J_2$ assigned to machine $M_1$ and $J_3$ to $M_2$ is in 1-1 correspondence with (ii) the \underline{non-}{conflict-free} extension
(in grey) in the corresponding AA framework 
where argument $a_{i, j}$ represents
assignment of $J_j$ to $M_i$ and attacks capture scheduling constraints. 
The (dashed) attack 
underlies an explanation as to why the schedule is inefficient: jobs $J_2$ and $J_3$ can be swapped between machines $M_2$ and $M_1$.
}
\label{fig:Scheduling}
\end{figure}

For  \textbf{planning}, 
\cite{Fan_18} generates an ABA counterpart of planning problems such that  extensions of the former correspond to  solutions to the latter. 
Also, \textbf{scheduling} has been targeted by
\cite{Cyras_19}, where a makespan scheduling problem with user decisions is translated into AA to 
extract sound and complete explanations as to why a schedule is (not) feasible, efficient and/or satisfies user decisions (see Figure~\ref{fig:Scheduling}).
Here, explanations 
are actionable in suggesting actions 
for fixing problems, if 
any exist.

Several complete post-hoc approaches have been proposed for {\bf logic programming}.
For example, \cite{Wakaki:Nitta:Sawamura:2009} show that exact mappings exist between answer sets 
and 
\AAALP, whereas
\cite{Schulz:Toni:2016}
explain
(non-)membership in answer sets in terms of 
ABA, based on sound and complete mappings between  answer sets and extensions
.
A similar approach is taken by \cite{Rolf_19}, who, 
however, use 
extensions for ADFs
.


\subsection{Approximate Post-Hoc Approaches}
\label{sec:sub:post-hoc-approximate}

\begin{figure}[t!]
\centering
\includegraphics[width=0.9\linewidth]{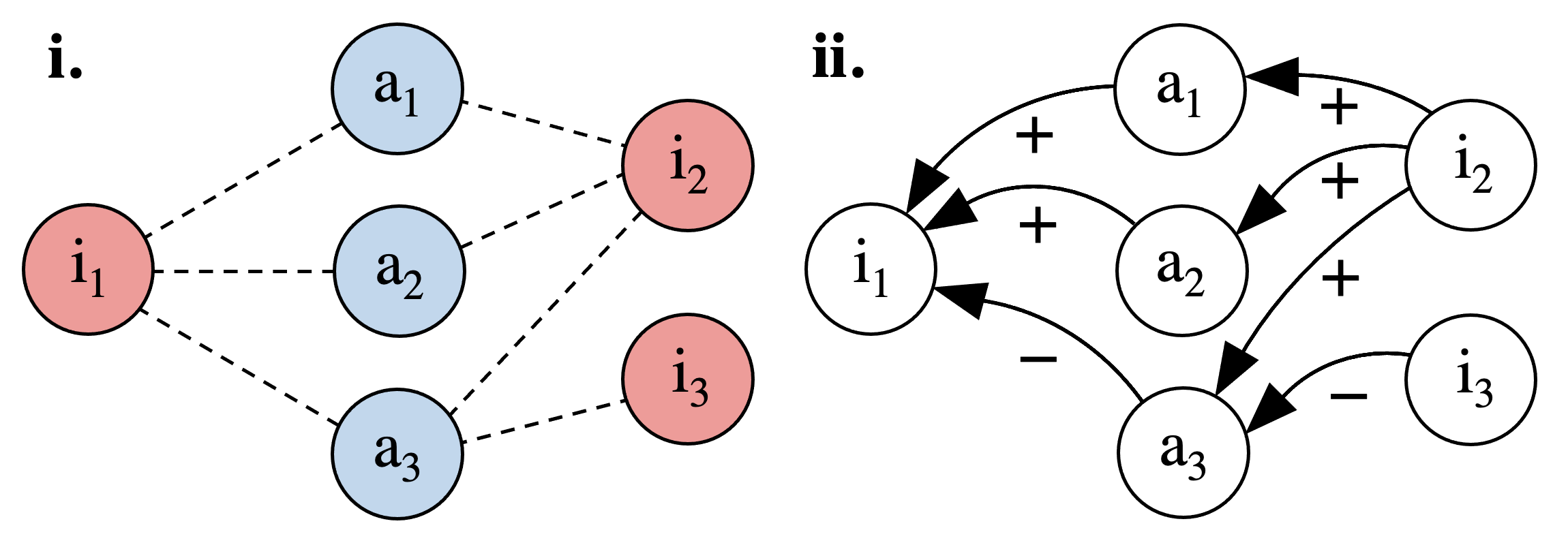}
\caption{Approximate post-hoc \argexpl s for 
RSs~\protect\cite{Rago_20}. 
(i) 
Connections between \underline{a}spects and \underline{i}tems 
underpinning the RS (predicted ratings omitted). (ii) Example BA framework extracted to explain 
the recommendation of 
$i_1$, where attacks (labelled~$-$) represent negative effects and supports (labelled~$+$) represent positive effects on predicted ratings.}
\label{fig:RS}
\end{figure}

Approximate post-hoc \argexpl s rely upon  incomplete mappings between the model to be explained and the AF from which explanations are drawn. 
For \textbf{RSs}, \cite{rec} use TA frameworks extracted from a hybrid RS as the basis for explanations. This RS exploits collaborative filtering and connections between items and their aspects 
to compute predicted ratings for items propagated through the connections. The TA framework approximates this propagation, 
categorising 
connections as dialectical relations 
if they lead to satisfaction of specified properties of 
TA semantics.
A similar RS with a greater explanatory repertoire, but supported by BA
, is given by \cite{Rago_20} (see Figure \ref{fig:RS}). 

Another category of \ARD{models }\ARF{model}
explained by approximated \argexpls{} are \textbf{classification} methods.  
\cite{Sendi.et.al:2019} 
extract rules from neural networks in an ensemble and deploy AA to explain their outputs.
Neural methods are also targeted by the general formalism of 
deep argumentative explanations
\cite{argflow}, which map neural networks to GA frameworks 
to explain their outputs in terms of 
inputs and, affording a \emph{deep} nature, 
intermediate components.

\begin{figure}[t!]
\centering
\includegraphics[width=1\linewidth]{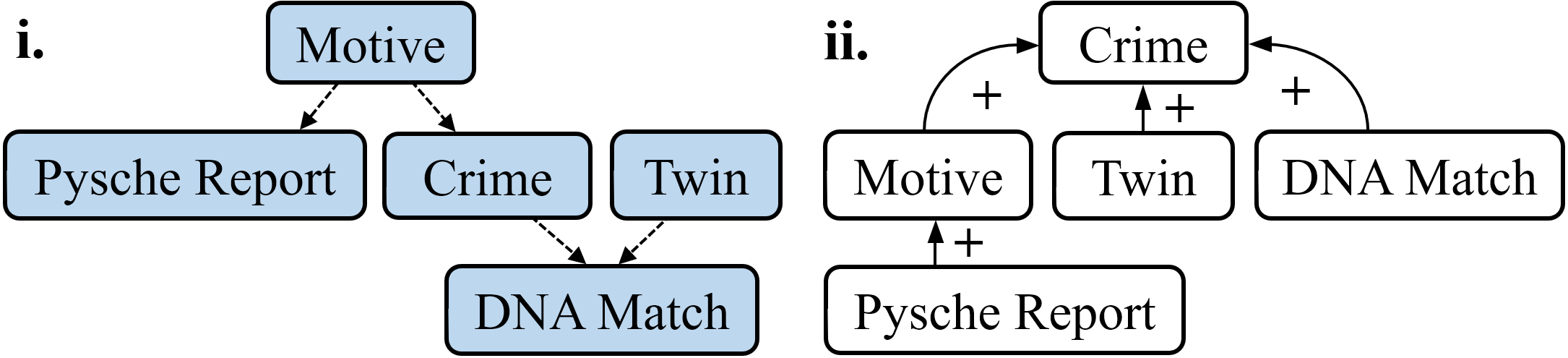}
\caption{Approximate post-hoc \argexpl s for Bayesian networks~\protect\cite{Timmer_17}. 
(i) Bayesian network with conditional dependencies 
(conditional probabilities ignored).
(ii) Extracted 
SA framework, where the support relation is directly derived from each variable's Markov blanket. 
}
\label{fig:BN}
\end{figure}

Approximated \argexpls{} have also been extracted from \textbf{probabilistic methods}, including Bayesian networks: for example, \cite{Timmer_17} model explanations as 
SA frameworks 
to show the interplay between variables in Bayesian networks (see Figure \ref{fig:BN}). 

Finally, to conclude our overview of approaches \ARD{to obtain }\ARF{for obtaining }\argexpl s,
\cite{Collins:Magazzeni:Parsons:2019} take  preliminary steps towards extracting approximate post-hoc \argexpls{} in \textbf{planning}: 
causal relationships are first extracted from a plan and then abstracted into 
ASPIC+
.

\section{Forms of \ARD{\Argexpls{}}\ARF{AF-Based Explanations}}
\label{sec:expl}
\label{sec:forms}

In §\ref{sec:applications} 
we discussed examples of three types of \argexpl{}. 
Here, we discuss various forms of argumentation-based structures that are used to formally define explanations in stand-alone AFs as well as AF-based agents. 
These could be (and often are) used in turn to support \argexpl s in their settings of use. 
We summarise the most representative such works in Table~\ref{table:forms} and illustrate the various forms of explanations by referring to Figures~\ref{fig:arg}--\ref{fig:BN}.

\begin{table}[h]
\begin{tabular}{lc}
& 
\!\!\!\!\!\!\!\!\!\!\!\!\!\!\!\textbf{Form of \ARD{e}\ARF{E}xplanation (AF)}
\\
\hline
\!\!\!\textbf{AFs} &
\\
\!\!\!\cite{Wakaki:Nitta:Sawamura:2009} & 
\!\!\!\!\!\!\!\!\!\!\!\!\!\!\!\dialogue{} (AA, \AAALP)
\\
\!\!\!\cite{Modgil:Caminada:2009} & 
\!\!\!\!\!\!\!\!\!\!\!\!\!\!\!\dialogue{} (AA)
\\
\!\!\!\cite{Seselja_13} &
\!\!\!\!\!\!\!\!\!\!\!\!\!\!\!Sub-graph (variant of AA)
\\
\!\!\!\cite{Grando_13} &
\!\!\!\!\!\!\!\!\!\!\!\!\!\!\!Sub-graph (ASPIC+)
\\
\!\!\!\cite{Garcia_13} & 
\!\!\!\!\!\!\!\!\!\!\!\!\!\!\!Sub-graph (DeLP)
\\
\!\!\!\cite{Booth.et.al:2014} & 
\!\!\!\!\!\!\!\!\!\!\!\!\!\!\!\!\!\!Change, \dialogue{} (AA, \AAALP)\!\!\!
\\
\!\!\!\cite{Fan_15} &
\!\!\!\!\!\!\!\!\!\!\!\!\!\!\!Extensions, DTs (AA, ABA)
\\
\!\!\!\cite{Fan:Toni:2015-TAFA} &
\!\!\!\!\!\!\!\!\!\!\!\!\!\!\!Change (AA)
\\
\!\!\!\cite{Cyras_19_ESA} &
\!\!\!\!\!\!\!\!\!\!\!\!\!\!\!DTs 
(\AACBR)
\\
\!\!\!\cite{Arioua_17} & 
\!\!\!\!\!\!\!\!\!\!\!\!\!\!\!\dialogue{} (\AALD) 
\\ 
\!\!\!\cite{Sakama:2018} & 
\!\!\!\!\!\!\!\!\!\!\!\!\!\!\!Change (AA)
\\
\!\!\!\cite{Zeng_19} & 
\!\!\!\!\!\!\!\!\!\!\!\!\!\!\!DTs (BA) 
\\
\!\!\!\cite{Saribatur.et.al:2020} &
\!\!\!\!\!\!\!\!\!\!\!\!\!\!\!Change (AA) 
\\ 
\!\!\!\cite{Liao_20} &
\!\!\!\!\!\!\!\!\!\!\!\!\!\!\!Extensions (AA)
\\
\hline
\!\!\!\textbf{AF-\ARD{b}\ARF{B}ased Agents} & 
\\
\!\!\!\cite{Gao_16} &
\!\!\!\!\!\!\!\!\!\!\!\!\!\!\!Extensions (AA) 
\\
\!\!\!\cite{Sklar_18} &
\!\!\!\!\!\!\!\!\!\!\!\!\!\!\!\dialogue{} (\AALD)
\\
\!\!\!\cite{Madumal_19} &
\!\!\!\!\!\!\!\!\!\!\!\!\!\!\!\dialogue{} (\AALD)
\\
\!\!\!\cite{Raymond:Gunes:Prorok:2020} &
\!\!\!\!\!\!\!\!\!\!\!\!\!\!\!\dialogue{} (\AALD)
\\ 
\!\!\!\cite{Sassoon_19} & 
\!\!\!\!\!\!\!\!\!\!\!\!\!\!\!\dialogue{} (\AAAS)
\\
\!\!\!\cite{Kokciyan_20} &
\!\!\!\!\!\!\!\!\!\!\!\!\!\!\!Structure 
(\AALD, ASPIC+)\!\!\!\!
\\
\!\!\!\cite{Espinoza_20} & 
\!\!\!\!\!\!\!\!\!\!\!\!\!\!\!Sub-graph, Extensions 
(\AALD)\!\!\!\!
\!\! \\
\hline
\end{tabular}
\caption{
Forms of explanations drawn from 
various forms of AF (stand-alone or in multi-agent contexts). 
}
\label{table:forms}
\end{table}

A common approach to explaining argument acceptability (i.e.\ membership in extensions) in AFs essentially amounts to traversing 
the AFs to show any arguments and 
elements of the dialectical relations relevant to determining a given argument's acceptance status. 
When the AFs can be understood as graphs (see Figure~\ref{fig:arg}i), this process amounts to identifying \textbf{sub-graphs} in the AFs, while also 
requiring 
satisfaction of some formal property 
(such as being a tree rooted in 
the argument whose acceptability status 
needs explaining, alternating attacking and defending arguments and with  
unattacked arguments as leaves). 
Formally, given a property $\prop$ of graphs and an AA framework $\graph = \langle \Args, \Atts \rangle$, an explanation for the acceptance status of some \emph{topic argument} $a\in \Args$ 
can be defined 
as
    a sub-graph $\graph' = \tuple{\Args', \mathcal{R}^{-'}}$ of $\graph$ such that $a \in \Args'$ and $\graph'$ satisfies $\prop$
\cite{Cocarascu.et.al:2018}.

A 
popular form of explanation as sub-graph is 
given by  \textbf{dispute trees}
(DTs, see §\ref{sec:AFs} and 
Figure~\ref{fig:arg}ii).
Importantly, 
DTs often carry theoretical guarantees 
towards desirable properties of explanations~\cite{Modgil:Caminada:2009,Fan_15,Cyras_19_ESA}, 
such as existence, correctness (the explanation actually establishes the acceptance status of the topic argument) and relevance (all arguments in the explanation 
play a role towards acceptance
)
. 

\argexpls{} as sub-graphs can also take forms of (sets of) paths, cycles or branches~\cite{Seselja_13,Garcia_13,Timmer_17,Cocarascu.et.al:2018,Cyras_19,Espinoza_20}. 
They are used 
with various AFs, 
e.g. BA~\cite{Cocarascu.et.al:2018,rec
}, QBA~\cite{RT} and 
DeLP~\cite{Garcia_13} 
(see Figures~\ref{fig:DeLP}--\ref{fig:BN} for examples).

Sub-graphs, especially DTs, act as proofs for argument acceptance~\cite{Modgil:Caminada:2009}. 
Similarly, in structured argumentation the (logic- or argument scheme-based) \textbf{structure} of arguments and dialectical relations 
can 
act as \ARD{explanations}\ARF{an explanation}~\cite{Moulin.et.al:2002,Grando_13,Naveed_18,Kokciyan_20}.

Whether comprising structured or abstract arguments, AFs naturally give rise to another form of \argexpl{}, namely a \textbf{dialogue} (\textbf{game}), whereby 
`participants engage in structured, rule-guided, goal-oriented exchange
' \cite{Sklar_18} of arguments in order to establish or explain argument acceptability. 
Dialogues can be constructed from DTs 
(see Figure~\ref{fig:arg}iii) or AFs in general \cite{Wakaki:Nitta:Sawamura:2009,Booth.et.al:2014,Arioua_17,Madumal_19,Raymond:Gunes:Prorok:2020,Walton:2004}, 
typically as formal games between two parties, \emph{proponent} and \emph{opponent}, with the aim of `winning' the game regarding the topic argument.
%

Explanations can also be \textbf{extensions} (see §\ref{sec:AFs}), 
keeping the relationships  implicit~\cite{Gao_16,Zeng_18,Espinoza_20,Liao_20}. 
For instance, \cite{Fan_15} define explanations as \emph{related admissible} 
extensions (conflict-free, defending against all attackers and in where one (topic) argument is defended by all other arguments) in AA frameworks. 
For illustration, in Figure~\ref{fig:arg}i, 
related admissible extensions $\{ a, c \}$ and $\{a, c, e\}$ are explanations of (the acceptability of) the topic argument $a$. 
%

Another form of 
\argexpl\ amounts to 
indicating 
a \textbf{change} of the AF
that would change some topic argument's acceptability status, often given some `universal' space of AF modifications~\cite{Wakaki:Nitta:Sawamura:2009,Booth.et.al:2014,Sakama:2018}. 
Specifically, addition or removal of arguments and/or relations that change the acceptability status of the topic argument is a form of explanation~\cite{Fan:Toni:2015-TAFA,Sakama:2018,Saribatur.et.al:2020}: 
in Figure~\ref{fig:arg}i, 
$\{ c \}$ can 
explain that removing $c$ from $\tuple{\Args, \Atts}$ would make $a$ non-acceptable (under, say, grounded extension semantics); 
likewise, 
\ARF{$(c,b) \in \Atts$}
can 
explain that removing the attack would make $a$ non-acceptable. 
%

Whichever the form of \argexpls{}, much work is often needed to  generate explanations presentable to the 
(human) users: this is the focus of several works, 
e.g.\ via natural language~\cite{Wyner_17,Madumal_19,Zhong_19,Cyras_19,RT,Raymond:Gunes:Prorok:2020,Espinoza_20} 
or visualisations~\cite{Grando_13,rec,Cyras_19_ESA,Zhong_19}. 
Popular are \emph{conversational} explanations~\cite{RT,Rago_20,Sklar_18,Sassoon_19,Madumal_19,Kokciyan_20}, 
often arising naturally from argument structure, sub-graphs and dialogue games (see Figure~\ref{fig:arg}iii).


\section{A Roadmap for Argumentative XAI}
\label{sec:roadmap}

Here we identify 
some gaps in the state-of-the-art on argumentation-based XAI 
and discuss opportunities for further research, focusing on three avenues: the need to devote more attention to \emph{properties} of \argexpls{}
; computational aspects of \argexpls{}
; and
broadening both applications and the scope of \argexpls{}. 

\subsubsection{Properties}
\label{sec:sub:prop}

Properties of AFs have been well studied, e.g.\ \cite{Dung:Mancarella:Toni:2007,Baroni_19}, but properties of \argexpls{} less so. 
Notable exceptions include forms of \emph{fidelity}, amounting to a sound and complete mapping from the system being explained and the generated \argexpls{}~\cite{Cyras_19,Fan_18}, and  
properties of extension-based \emph{explanation semantics}~\cite{Liao_20}. 
Other desirable properties from the broader XAI landscape~\cite{Sokol_20} have been mostly neglected, 
though some user-acceptance aspects such as
\emph{cognitive tractability}~\cite{Cyras_19}
as well as \emph{transparency} and \emph{trust}~\cite{Rago_20} have been considered for \argexpls{}. 
For some properties, experiments with human users may be needed, as in much of the XAI literature~\cite{Adadi:Berrada:2018}, 
and creativity in the actual format of \argexpls{} shown to humans required.

\subsubsection{Computational Aspects}
\label{sec:sub:computation}

To effectively support XAI solutions, \argexpls{} need to be
efficiently computable. 
In the case of \emph{intrinsic \argexpls{}}, this requires efficient systems for the relevant reasoning tasks
and 
a good understanding of the computational complexity 
thereof.
For illustration, the approaches of \cite{Cyras_19_ESA,Cocarascu_20} rely upon the tractable membership reasoning task 
for the grounded extension for AA.
In the case of \emph{post-hoc (complete or approximate)} \argexpl s, a further hurdle is the extraction of AFs from the models in need of explanation, prior to the extraction of the \argexpl s themselves. 
For illustration, the (complete) approach of 
\cite{Cyras_19} proves `soundness and completeness' tractability. 

For all types of \argexpl s, further consideration must be given to the extraction task, of explanations of various formats from AFs. 
For illustration, the \argexpl s for the approaches of \cite{Cyras_19_ESA,Cocarascu_20} rely upon DTs that can be extracted efficiently from 
AFs, given the grounded extension. 
Further,
\cite{Saribatur.et.al:2020} give complexity results for extracting certain sets of arguments as explanations in AA.  
In general, however, computational issues in \argexpl s require a more systematic investigation both in terms of underpinning reasoning tasks and explanation extraction.

\subsubsection{Extending applications and the scope of explanations}
\label{sec:sub:applications}

While already having a variety of instantiations and covering a wide range of application contexts, AF-based explanations have a wide potential of further development. 

Concerning applications, arguably the strongest demand for XAI solutions is currently driven by applications of machine learning (ML). 
In this context, it is interesting to note that in a loose sense some forms of ML have dialectical roots: supervised ML uses positive and negative examples of concepts to be learnt, and reinforcement learning uses positive and negative rewards. 
Further, several of the existing XAI solutions for ML, albeit not explicitly argumentative in the sense of this survey, 
are
argumentative in spirit, as discussed in §\ref{sec:Intro} (e.g.\ SHAP~\cite{Lundberg_17} can be seen as identifying reasons for and against outputs). 
However, \argexpls{} have been only sparingly deployed in ML-driven (classification and probabilistic) settings (see Table~\ref{table:applications}). 
We
envisage a fruitful interplay, where the explanation needs of ML, while benefiting from the potential of argumentation techniques, also stimulate further research 
in computational argumentation. 
Specifically, at the ML end, 
the analysis of dialectics is a crucial, yet often ignored, underpinning of XAI for ML: it would then be interesting to explore whether an understanding of some existing methods in terms of \argexpl s could pave the way to new developments. This in turn could lead, on the computational argumentation side, to novel forms of explanation-oriented AFs.
 
As a first step, it would be interesting to see whether existing approaches on logic-based explanations, either model-agnostic \cite{Ignatiev_19,Darwiche_20} or model-specific \cite{Shih_18}, could be understood as \argexpl s, potentially relying upon existing logic-based AFs such as \cite{Besnard_01}, or ADFs/AFs with structured arguments (see §\ref{sec:AFs}).
Connections with the widely used \emph{counterfactual explanations (CFs)} (e.g.\ see \cite{Sokol_20}) represent another stimulating investigation topic.
CFs identify, as explanations for models' outputs, hypothetical changes in the inputs that would change these outputs. As such, they show, again, some dialectical flavour and call for the study of forms of \argexpl s able to provide CF functionalities. 
For instance, 
\emph{relation-based CFs}~\cite{CFX} may be interpretable in terms of \argexpl s for suitable AFs (accommodating different types of support to match the underpinning relations). 
Given that CFs are based on `changes', the corresponding form of \argexpls{} discussed in §\ref{sec:forms} could support this kind of development also.


\section{Conclusions}
\label{sec:conclusions}

We have given a comprehensive survey of the 
active research area of argumentative XAI, focusing on explanations built using argumentation frameworks (AFs) from computational argumentation.  
We have shown how varied \argexpls{} are, in terms of the models they explain, the AFs they deploy, and the forms they take. 
We have also discussed how \argexpl s may be defined as integral components of systems that are (argumentatively) explainable by design (we called these explanations `intrinsic'), but, in several settings, may be provided for existing systems in need of explaining (we called these explanations `post-hoc') as a result of a marriage between symbolic representations (the AFs) and various forms of (e.g.\ symbolic or statistical/probabilistic) models.
Finally, we have set out a roadmap for future developments of \argexpl s and their use, which we hope will be beneficial to the AI research community at large, from experts in symbolic AI (and computational argumentation in particular) to application experts in need of customisable, powerful XAI solutions.

\section*{Acknowledgements}  

This research was funded in part by the Royal Academy of Engineering, UK, and by J.P. Morgan.

\small{
\bibliographystyle{named}
\bibliography{bib-noproc}
}

\end{document}